\documentclass{article}
\usepackage{graphicx} 
\usepackage{authblk}
\usepackage{amsmath}
\usepackage{amssymb}
\usepackage{hyperref}
\usepackage[labelfont=bf,textfont=md]{caption}
\usepackage{tikz}
\usepackage{pgfplots}

\usepackage{setspace}
\let\svthefootnote\thefootnote
\newcommand\freefootnote[1]{%
  \let\thefootnote\relax%
  \footnotetext{#1}%
  \let\thefootnote\svthefootnote%
}

\title{Leveraging Language Models and RAG for Efficient Knowledge Discovery in Clinical Environments}
\author[1]{Seokhwan Ko}
\author[2]{Donghyeon Lee}
\author[2]{Jaewoo Chun}
\author[1,3]{Hyungsoo Han}
\author[1,*]{Junghwan Cho}

\affil[1]{Clinical Omics Institute, Kyungpook National University}
\affil[2]{Department of Biomedical Science, School of Medicine Kyungpook National University}
\affil[3]{Department of Physiology, School of Medicine Kyungpook National University}

\date{}

\begin{document}

\maketitle

\freefootnote{AI Transformation Challenge and Symposium 2025}
\freefootnote{* Corresponding author, joshua@knu.ac.kr}
\begin{abstract}
Large language models (LLMs) are increasingly recognized as valuable tools across the medical environment, supporting clinical, research, and administrative workflows. However, strict privacy and network security regulations in hospital settings require that sensitive data be processed within fully local infrastructures. Within this context, we developed and evaluated a retrieval-augmented generation (RAG) system designed to recommend research collaborators based on PubMed publications authored by members of a medical institution. The system utilizes PubMedBERT for domain-specific embedding generation and a locally deployed LLaMA3 model for generative synthesis. This study demonstrates the feasibility and utility of integrating domain-specialized encoders with lightweight LLMs to support biomedical knowledge discovery under local deployment constraints.

\end{abstract}

\section{Introduction}

The growing complexity of clinical and biomedical information has increased the demand for tools that can support interpretation, reporting, and information retrieval across medical and academic environments. LLM-driven systems have emerged as effective solutions, enabling automated document generation, structured reporting, and administrative support \cite{Singhal2023large, zhou2025large}. They further accelerate literature review and facilitate collaboration discovery by synthesizing large volumes of biomedical text \cite{nori2023capabilities, tang2025large}.

However, in hospital environments, regulatory constraints require that patient-sensitive or institution-specific data remain within secure, isolated networks \cite{yigzaw2022health,khalid2023privacy,kelly2023cybersecurity}. This limits the use of cloud-based AI services and motivates the development of locally deployable LLM systems \cite{basit2024medaide}.

In this preliminary study, we present a research collaboration recommendation system designed for institutional deployment. The system leverages PubMed publication metadata and generative modeling to identify potential collaborators, summarize research topics, and facilitate interdisciplinary discovery across the Kyungpook National University (KNU) School of Medicine.

\section{Materials and Methods}

\begin{figure*}[h]
\centering
\includegraphics[width=\textwidth]{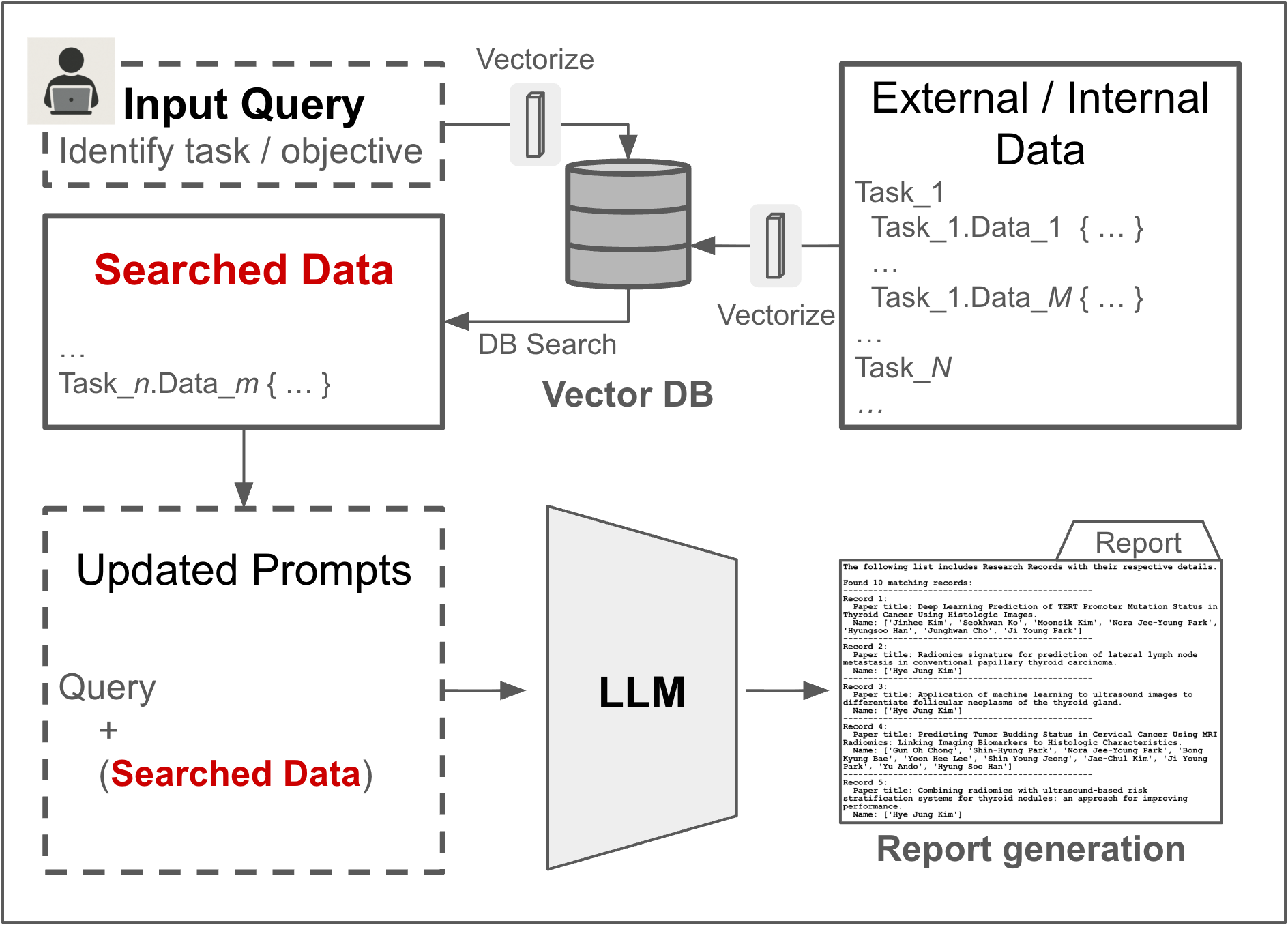}
\caption{Overall workflow of the collaboration recommendation system, based on a Retrieval-Augmented Generation (RAG) architecture.}
\label{fig1}
\end{figure*}

To construct the institutional knowledge base, publication records authored by researchers affiliated with the KNU School of Medicine were collected from PubMed \cite{lu2011pubmed, wei2019pubtator}. For each entry, metadata such as titles, abstracts, author lists, affiliations, keywords, and publication years were extracted. All documents were stored locally within the hospital network to satisfy data security and privacy requirements. This curated set of structured publication records served as the foundation for subsequent embedding and retrieval processes. An overview of the entire workflow is shown in Figure~\ref{fig1}.

\subsection{Embedding Representation}

We represent the entire corpus of PubMed abstracts as
$\mathcal{D}=\{d_1, d_2, \dots, d_N\}$. Each document is encoded into a dense biomedical semantic embedding using PubMedBERT \cite{gu2021domain}:

\begin{equation}
\mathbf{h}_i = f_{\text{PB}}(d_i) \in \mathbb{R}^m,
\end{equation}

where $f_{\text{PB}}(\cdot)$ denotes the domain-specific encoder and $m$ is the embedding dimension.  
A user query $q$ is processed in the same manner:

\begin{equation}
\mathbf{h}_q = f_{\text{PB}}(q).
\end{equation}

All embeddings $\{\mathbf{h}_i\}$ were indexed in a local vector database to enable efficient semantic retrieval, following standard practices in approximate nearest-neighbor search \cite{johnson2019billion}.

\subsection{Semantic Retrieval Using Cosine Similarity}

To identify publications most relevant to a user query, we compute cosine similarity between the query embedding and each document embedding, a widely used metric in vector-space retrieval \cite{schutze2008introduction, reimers2019sentence}:

\begin{equation}
\text{sim}(\mathbf{h}_q, \mathbf{h}_i) =
\frac{\mathbf{h}_q \cdot \mathbf{h}_i}
{\|\mathbf{h}_q\| \, \|\mathbf{h}_i\|}.
\end{equation}

Documents are ranked according to their similarity scores, and the system selects the top-$K$ results:

\begin{equation}
\mathcal{R}_K(q) = 
\operatorname{TopK}_{d_i \in \mathcal{D}} \; \text{sim}(\mathbf{h}_q, \mathbf{h}_i).
\end{equation}

This retrieval step ensures that the generative model receives both the user’s intent and a set of semantically aligned evidence from the literature.

\subsection{Prompt Construction for RAG}

The retrieved documents are incorporated into a retrieval-augmented prompt that provides contextual grounding for the generative model, in line with standard RAG approaches \cite{lewis2020retrieval, izacard2021leveraging}. This prompt is constructed by concatenating the original query with the selected documents:

\begin{equation}
P(q) = \operatorname{Concat}\!\left(q,\; d_{(1)}, d_{(2)}, \dots, d_{(K)}\right),
\end{equation}

where $d_{(j)}$ denotes the $j$-th highest-ranked retrieved document according to cosine similarity, ensuring that the prompt preserves the relevance-based ordering of retrieved contexts.

\subsection{Generative Synthesis With LLaMA3.2}

To comply with network security policies within the hospital environment, all generative inference is performed locally using LLaMA3.2, a 3B-parameter lightweight model \cite{touvron2023llama}. Given the retrieval-augmented prompt $P(q)$, the model synthesizes summary information and produces a ranked recommendation of potential collaborators:

\begin{equation}
y = g_{\text{LLM}}(P(q)),
\end{equation}

where $g_{\text{LLM}}$ denotes the generative model. The output combines information inferred from the query, retrieved publications, and patterns captured during pretraining, resulting in interpretable recommendations and topic summaries aligned with the institution’s research landscape.

\section{Results}

\begin{figure*}[ht]
\centering
\includegraphics[width=\textwidth]{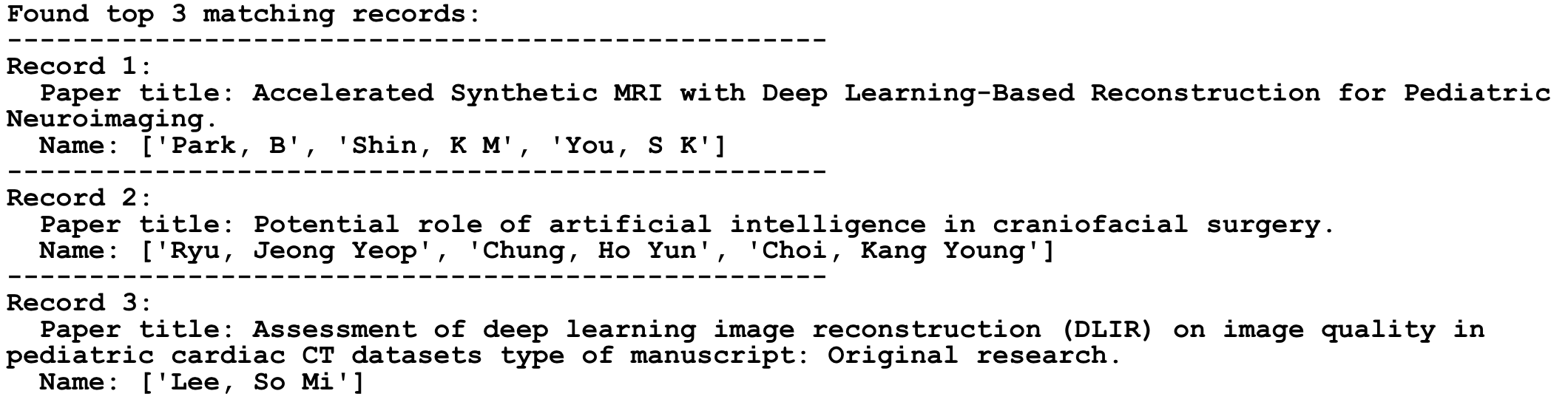}
\caption{Example output for the query ``deep learning prediction for medical images.'' Recommended researchers and topics are synthesized from retrieved PubMed publications.}
\label{fig2}
\end{figure*}

Pilot evaluations showed that the system effectively retrieved contextually relevant publications and synthesized informative collaboration suggestions. For the query \textit{“deep learning prediction for medical images”}, the system identified research groups specializing in thyroid pathology, deep learning, medical imaging, and endocrine oncology, as shown in Figure~\ref{fig2}.

Compared with traditional keyword-based PubMed searches, PubMedBERT-based embeddings improved contextual retrieval quality, and LLaMA3.2 provided concise, interpretable summaries that highlighted research themes, methodologies, and potential interdisciplinary links.

\begin{figure*}[ht]
\centering
\includegraphics[width=\textwidth]{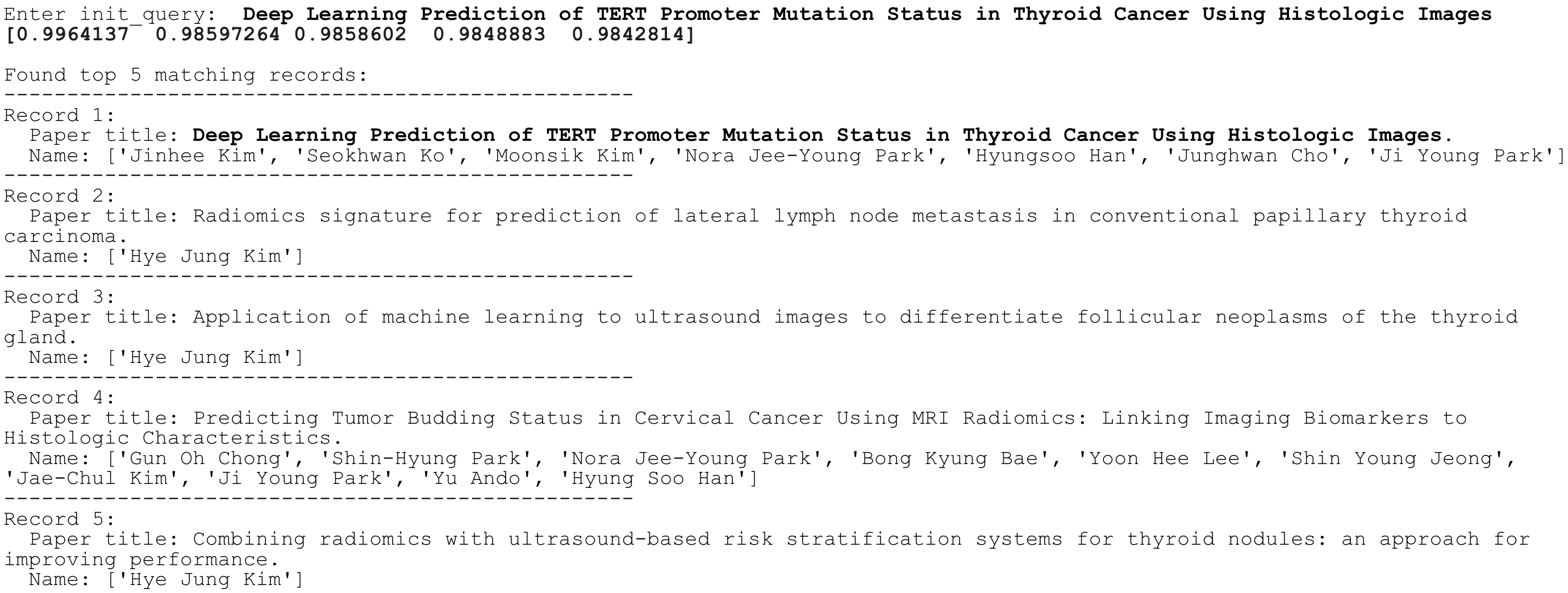}
\caption{Cosine similarity scores for the top retrieved documents given the query 
``Deep Learning Prediction of TERT Promoter Mutation Status in Thyroid Cancer Using Histologic Images.'' 
The exact matching publication is ranked first with the highest similarity score, confirming correct retrieval behavior.}
\label{fig3}
\end{figure*}

To validate the correctness of the embedding and similarity computation, we examined the cosine similarity scores for a representative query:
\textit{``Deep Learning Prediction of TERT Promoter Mutation Status in Thyroid Cancer Using Histologic Images.''}
As expected, the system returned the exact matching publication as the top-ranked result with a similarity score of $0.9964137$.
The subsequent retrieved documents showed cosine similarity scores of $0.9859726$, $0.9858602$, $0.9848883$, and $0.9842814$ respectively.
Figure~\ref{fig3} visualizes these similarity scores and confirms that the cosine similarity module operates as intended.

\section{Conclusion}

We present a locally deployable RAG system that integrates PubMedBERT for semantic retrieval and LLaMA3.2 for generative analysis. The system enhances research networking efficiency while adhering to strict privacy and security constraints in hospital environments. This approach demonstrates the feasibility of deploying lightweight yet domain-effective LLM systems for biomedical knowledge discovery.

\section{Discussion}

This study demonstrates that integrating domain-specialized encoders \cite{gu2021domain} with a lightweight locally deployed LLM can yield an effective and interpretable system for identifying potential research collaborations within a medical institution. By leveraging PubMed-derived publication metadata and retrieval-augmented generation \cite{lewis2020retrieval}, the framework provides structured, context-aware recommendations while remaining compatible with strict security and deployment constraints in hospital environments.

The system also opens several avenues for further enhancement. One promising direction is the incorporation of agentic components capable of autonomously monitoring newly published literature \cite{wang2024survey}, thereby enabling continuous updates and real-time detection of emerging research themes \cite{liu2023llm}. Extending the framework to support cross-institutional biomedical knowledge graph construction \cite{nicholson2020constructing} would further enrich representations of collaborative networks and scientific domains. Moreover, connecting the system’s outputs to institutional grant management pipelines and research workflow automation tools could facilitate more seamless integration into operational research environments, ultimately supporting strategic planning and interdisciplinary collaboration at scale.

\section{Acknowlegment}
This research was supported by the Brain Pool Program through the National Research Foundation of Korea (NRF), funded by the Ministry of Science and ICT (Grant No. 2022H1D3A2A01096490 \& RS-2023-00283791) and the Ministry of Education, Korea (Grant No. 2021R1I1A3056903 \& RS-2024-00459836).
Special thanks to Yu Ando, for his insightful comments and discussions.

\bibliographystyle{ieeetr}
\bibliography{ref}

@article{singhal2023large,
  title={Large language models encode clinical knowledge},
  author={Singhal, Karan and Azizi, Shekoofeh and Tu, Tao and Mahdavi, S Sara and Wei, Jason and Chung, Hyung Won and Scales, Nathan and Tanwani, Ajay and Cole-Lewis, Heather and Pfohl, Stephen and others},
  journal={Nature},
  volume={620},
  number={7972},
  pages={172--180},
  year={2023},
  publisher={Nature Publishing Group}
}

@article{zhou2025large,
  title={Large language models in biomedicine and healthcare},
  author={Zhou, Juexiao and Li, Haoyang and Chen, Siyuan and Chen, Zhangtianyi and Han, Zhongyi and Gao, Xin},
  journal={npj Artificial Intelligence},
  volume={1},
  number={1},
  pages={44},
  year={2025},
  publisher={Nature Publishing Group UK London}
}

@article{nori2023capabilities,
  title={Capabilities of gpt-4 on medical challenge problems},
  author={Nori, Harsha and King, Nicholas and McKinney, Scott Mayer and Carignan, Dean and Horvitz, Eric},
  journal={arXiv preprint arXiv:2303.13375},
  year={2023}
}

@inproceedings{tang2025large,
  title={Large language models for automated literature review: An evaluation of reference generation, abstract writing, and review composition},
  author={Tang, Xuemei and Duan, Xufeng and Cai, Zhenguang},
  booktitle={Proceedings of the 2025 Conference on Empirical Methods in Natural Language Processing},
  pages={1602--1617},
  year={2025}
}

@article{yigzaw2022health,
  title={Health data security and privacy: Challenges and solutions for the future},
  author={Yigzaw, Kassaye Yitbarek and Olabarriaga, S{\'\i}lvia Delgado and Michalas, Antonis and Marco-Ruiz, Luis and Hillen, Christiaan and Verginadis, Yiannis and De Oliveira, Marcela Tuler and Krefting, Dagmar and Penzel, Thomas and Bowden, James and others},
  journal={Roadmap to successful digital health ecosystems},
  pages={335--362},
  year={2022},
  publisher={Elsevier}
}

@article{khalid2023privacy,
  title={Privacy-preserving artificial intelligence in healthcare: Techniques and applications},
  author={Khalid, Nazish and Qayyum, Adnan and Bilal, Muhammad and Al-Fuqaha, Ala and Qadir, Junaid},
  journal={Computers in Biology and Medicine},
  volume={158},
  pages={106848},
  year={2023},
  publisher={Elsevier}
}

@article{kelly2023cybersecurity,
  title={Cybersecurity considerations for radiology departments involved with artificial intelligence},
  author={Kelly, Brendan S and Quinn, Conor and Belton, Niamh and Lawlor, Aonghus and Killeen, Ronan P and Burrell, James},
  journal={European radiology},
  volume={33},
  number={12},
  pages={8833--8841},
  year={2023},
  publisher={Springer}
}

@article{gu2021domain,
  title={Domain-specific language model pretraining for biomedical natural language processing},
  author={Gu, Yu and Tinn, Robert and Cheng, Hao and Lucas, Michael and Usuyama, Naoto and Liu, Xiaodong and Naumann, Tristan and Gao, Jianfeng and Poon, Hoifung},
  journal={ACM Transactions on Computing for Healthcare (HEALTH)},
  volume={3},
  number={1},
  pages={1--23},
  year={2021},
  publisher={ACM New York, NY}
}

@article{touvron2023llama,
  title={Llama: Open and efficient foundation language models},
  author={Touvron, Hugo and Lavril, Thibaut and Izacard, Gautier and Martinet, Xavier and Lachaux, Marie-Anne and Lacroix, Timoth{\'e}e and Rozi{\`e}re, Baptiste and Goyal, Naman and Hambro, Eric and Azhar, Faisal and others},
  journal={arXiv preprint arXiv:2302.13971},
  year={2023}
}

@article{basit2024medaide,
  title={MedAide: leveraging large language models for on-premise medical assistance on edge devices},
  author={Basit, Abdul and Hussain, Khizar and Hanif, Muhammad Abdullah and Shafique, Muhammad},
  journal={arXiv preprint arXiv:2403.00830},
  year={2024}
}

@article{lu2011pubmed,
  title={PubMed and beyond: a survey of web tools for searching biomedical literature},
  author={Lu, Zhiyong},
  journal={Database},
  volume={2011},
  pages={baq036},
  year={2011},
  publisher={Oxford University Press}
}

@article{wei2019pubtator,
  title={PubTator central: automated concept annotation for biomedical full text articles},
  author={Wei, Chih-Hsuan and Allot, Alexis and Leaman, Robert and Lu, Zhiyong},
  journal={Nucleic acids research},
  volume={47},
  number={W1},
  pages={W587--W593},
  year={2019},
  publisher={Oxford University Press}
}

@book{schutze2008introduction,
  title={Introduction to information retrieval},
  author={Sch{\"u}tze, Hinrich and Manning, Christopher D and Raghavan, Prabhakar},
  volume={39},
  year={2008},
  publisher={Cambridge University Press Cambridge}
}

@article{reimers2019sentence,
  title={Sentence-bert: Sentence embeddings using siamese bert-networks},
  author={Reimers, Nils and Gurevych, Iryna},
  journal={arXiv preprint arXiv:1908.10084},
  year={2019}
}

@article{johnson2019billion,
  title={Billion-scale similarity search with GPUs},
  author={Johnson, Jeff and Douze, Matthijs and J{\'e}gou, Herv{\'e}},
  journal={IEEE Transactions on Big Data},
  volume={7},
  number={3},
  pages={535--547},
  year={2019},
  publisher={IEEE}
}

@article{lewis2020retrieval,
  title={Retrieval-augmented generation for knowledge-intensive nlp tasks},
  author={Lewis, Patrick and Perez, Ethan and Piktus, Aleksandra and Petroni, Fabio and Karpukhin, Vladimir and Goyal, Naman and K{\"u}ttler, Heinrich and Lewis, Mike and Yih, Wen-tau and Rockt{\"a}schel, Tim and others},
  journal={Advances in neural information processing systems},
  volume={33},
  pages={9459--9474},
  year={2020}
}

@inproceedings{izacard2021leveraging,
  title={Leveraging passage retrieval with generative models for open domain question answering},
  author={Izacard, Gautier and Grave, Edouard},
  booktitle={Proceedings of the 16th conference of the european chapter of the association for computational linguistics: main volume},
  pages={874--880},
  year={2021}
}

@article{wang2024survey,
  title={A survey on large language model based autonomous agents},
  author={Wang, Lei and Ma, Chen and Feng, Xueyang and Zhang, Zeyu and Yang, Hao and Zhang, Jingsen and Chen, Zhiyuan and Tang, Jiakai and Chen, Xu and Lin, Yankai and others},
  journal={Frontiers of Computer Science},
  volume={18},
  number={6},
  pages={186345},
  year={2024},
  publisher={Springer}
}

@article{liu2023llm,
  title={Llm-powered hierarchical language agent for real-time human-ai coordination},
  author={Liu, Jijia and Yu, Chao and Gao, Jiaxuan and Xie, Yuqing and Liao, Qingmin and Wu, Yi and Wang, Yu},
  journal={arXiv preprint arXiv:2312.15224},
  year={2023}
}

@article{nicholson2020constructing,
  title={Constructing knowledge graphs and their biomedical applications},
  author={Nicholson, David N and Greene, Casey S},
  journal={Computational and structural biotechnology journal},
  volume={18},
  pages={1414--1428},
  year={2020},
  publisher={Elsevier}
}

\end{document}